# Agrupamento de Pixels para o Reconhecimento de Faces[*]

Tiago Buarque Assunção de Carvalho

10 de junho de 2020

***Abstract***

This research starts with the observation that face recognition can suffer a low impact from significant image shrinkage. To explain this fact, we proposed the Pixel Clustering methodology. It defines regions in the image in which its pixels are very similar to each other. We extract features from each region. We used three face databases in the experiments. We noticed that 512 is the maximum number of features needed for high accuracy image recognition. The proposed method is also robust, even if only it uses a few classes from the training set.

***Resumo***

Esta pesquisa começa com a observação de que o reconhecimento facial pode sofrer um baixo impacto devido ao encolhimento significativo da imagem. Para explicar esse fato, propusemos a metodologia Agrupamento de Pixels. Esta define regiões na imagem em que seus pixels são muito semelhantes entre si. Extraímos características de cada região. Utilizamos três bancos de dados de face nos experimentos. Percebemos que 512 é o número máximo de características necessárias para o reconhecimento de imagens com alta acurácia. O método proposto também é robusto, mesmo que use apenas algumas classes do conjunto de treinamento.

## 1 Introdução

### 1.1 Visão Computacional

Visão Computacional (VC) é uma área da Ciência da Computação que está entre duas grandes áreas: Aprendizagem de Máquina (Reconhecimento de Padrões, Redes Neurais, Inteligência Artificial etc.) e Processamento Digital de Imagens (Computação Gráfica etc.). A Aprendizagem de Máquina (AM) está também ligada a outras área da Computação como Mineração de Dados, por exemplo. O objetivo da AM é dar sentido aos dados, isto é muito valioso na era da *Big Data* que vivemos. AM também é utilizada para filtrar *spams* em e-mail, ou para detectar padrões inseguros em redes de computador.

A VC é a chave para muitas aplicações recentes, como carros autônomos. E também é componente importante em sistemas que envolvem outras áreas da Computação. Utiliza-se VC para ligar dispositivos móveis com sistemas web através de *QR codes*. Para prover mais segurança a sistemas utiliza-se reconhecimento face, de impressões digitais ou iris. Alguns bancos de dados de fotos organizam suas imagens a partir das faces das pessoas e até de animais. O reconhecimento de faces, como visto, é uma área da VC com várias aplicações na Computação.

---

[*]Este artigo foi submetido como capítulo de livro organizado pelo curso de Bacharelado em Ciência da Computação da Universidade Federal do Agreste de Pernambuco.

## 1.2 Reconhecimento de Faces

Reconhecimento de faces é uma tarefa trivial ao ser humano, mas tem sido um desafio para a visão computacional. O reconhecimento pode se dar a partir de três fontes distintas de imagens: vídeos, imagens estáticas 2D e imagens 3D (Zhao et al. 2003). Nesta pesquisa são abordadas apenas imagens estáticas 2D. Exemplos deste tipo de imagem são fotografias de faces. Embora muitas das informações nos parágrafos seguintes também possam ser aplicadas aos outros tipos de imagens de face, a discussão restringe-se a imagens estáticas 2D.

Um sistema genérico de reconhecimento de faces possui três principais etapas: detecção da face, extração de características e reconhecimento (Zhao et al. 2003). A detecção de faces consiste em encontrar em qual região da imagem está a face. A extração de características é a etapa que processa a imagem em busca de informações representativas e úteis para classificá-las. Esta etapa depende fortemente da aplicação. Por exemplo, características para reconhecer uma pessoa da foto podem não ser tão úteis para identificar a expressão da face. A etapa de reconhecimento é aquela na qual se utiliza um classificador para realizar ou a tarefa de verificação ou identificação. Huang et al. (Huang et al. 2008) propõem, para a identificação de faces, a sequência: detecção, alinhamento e reconhecimento. Antes de realizar o reconhecimento, detecta-se a posição da face na imagem e alinha-se a face.

Cada um destes elementos (detecção, extração de características, reconhecimento) tem um papel importante em um sistema de reconhecimento de faces. Como realizado pelos autores comentados acima (Huang et al. 2008; Simonyan et al. 2013) e por outros (Zhao et al. 2003; Jones 2009), as pesquisas focam em resolver problemas em apenas uma etapa. O foco desta pesquisa é a extração de características. Para tanto assume que estas etapas de pré-processamento estão resolvidas: a imagem recebida como entrada está devidamente recortada e alinhada. E na etapa de reconhecimento utiliza classificadores propostos por outros autores.

Jones (Jones 2009) analisa vários artigos desenvolvidos em reconhecimento de faces e enumera os problemas desta tarefa: iluminação irregular na foto, pose (rotação da cabeça), expressão do rosto, envelhecimento/crescimento da pessoa, alinhamento impreciso. Dentre estes problemas a pose é um problema bastante severo, pois a imagem do perfil de uma face é bem diferente da imagem frontal da mesma face, uma abordagem para corrigir este problema é proposta por Yi et al. (Yi, Lei e Li 2013), um modelo para mapear imagens 2D em um modelo 3D de face. Na presente pesquisa também é assumido que as imagens não apresentam problemas severos de pose e alinhamento, e que as técnicas propostas poderiam ser utilizadas em qualquer base se estes problemas fossem corrigidos com algum pré-processamento como o proposto por Yi et al.

## 1.3 Dados de alta dimensionalidade

É válido enfatizar que as imagens de face utilizadas aqui são: estáticas, 2D, em tons de cinza, recortadas e contendo apenas a região da face, bem centralizadas, sem sérios problemas de pose e rotação da cabeça, contendo algum problema de iluminação. Assume-se a representação de uma imagem como uma matriz onde cada posição da matriz é denominada pixel e tem seu valor em um intervalo, geralmente [0, 255]. Com estas restrições, as imagens de face são dados de alta dimensionalidade em que muitas variáveis são correlacionadas. Os dados têm alta dimensionalidade porque cada pixel da imagem é considerado uma característica e mesmo para imagens pequenas o número de pixels é muito alto, por exemplo, uma imagem $100 \times 100$ tem 10.000 características.

A expressão alta dimensionalidade também é empregada quando o número de características é muito maior que o número de exemplos de treino (Hastie, Tibshirani e Friedman 2001), o que também ocorre com as imagens de face. Esses dados têm muitas características correlacionadas porque várias regiões da imagem têm intensidade de pixels semelhantes, por exemplo os pixels na região de: pelo, cabelo, olhos, barba etc. Desta maneira a extração de características tem o objetivo não apenas de gerar novas características discriminantes (capaz de melhor separar as classes), como também reduzir a dimensionalidade do problema, por exemplo, de 10.000 para 40 características (cada característica é interpretada como uma dimensão).

Pode-se assumir que os pixels de cada imagem representam regiões correspondentes. Por exemplo, um pixel que representa uma região na bochecha de uma pessoa deve representar também a região da bochecha na maioria das imagens. Com as faces alinhadas desta maneira, faz sentido compará-las através de uma medida de distância, como a distância Euclidiana, por exemplo. Desta maneira, também restringem-se os algoritmos de classificação utilizados. Utiliza-se principalmente o classificador pelo vizinho mais próximo (1-NN, 1-*Nearest Neighbor*), o qual atribui para uma nova imagem de face a mesma classe da imagem mais próxima de acordo com a distância Euclidiana.

## 1.4 Extração de Características

Uma vez delimitados os demais elementos do sistema de reconhecimento de faces, são propostos algoritmos de extração de características para reduzir as dimensões destes dados de alta dimensionalidade. Tais algoritmos são não-supervisionados, isto é, não dependem dos rótulos das classes das amostras de treino. Um dos grupos de algoritmos propostos foram inspirados no método *Waveletfaces* (Chien e Wu 2002). Após uma análise detalhada deste método percebeu-se que a simples redução da imagem funciona muito bem como extração de característica, não diminuindo o erro de classificação até um certo limite de redução.

Como reduzir a imagem não impacta a taxa de reconhecimento de faces? Na tentativa de explicar esse fenômeno é proposta nesta pesquisa a metodologia de Agrupamento de Pixels. Esta metodologia coloca em um mesmo grupo os pixels com intensidade semelhante em todas as imagens do conjunto de treinamento. Cada grupo define uma região cuja média em cada imagem é uma característica extraída. Na literatura foram encontra-dos trabalhos semelhantes à proposta de agrupamento de pixels, chamadas Agrupamento de Características. A grande maioria deste trabalhos foca na tarefa de reconhecimento de texto, apenas os trabalhos de Avidan et al. (Shai Avidan 2002; S. Avidan e Butman 2004) e Song et al. (Song, Ni e Wang 2013) utilizam bases de dados de faces. Apenas o trabalho *Eigensegments* (Shai Avidan 2002) de Avidan aborda diretamente o problema de identificação de faces. Diferentemente do método proposto nesta pesquisa, Avidan extrai características utilizando utilizando *Eigenfaces* (Turk e Pentland 1991).

## 1.5 Objetivos

O número muito alto de dimensões se torna um fator limitante para a tarefa de classificação. Este fenômeno é conhecido como maldição da dimensionalidade. Reduzir a dimensionalidade dos dados é uma forma de aumentar a taxa de classificações corretas. O objetivo geral nesta pesquisa é propor uma nova metodologia de extração de características, para redução de dimensionalidade, baseada em Agrupamento de Características.

Os objetivos específicos são:

- restringir a pesquisa a problemas com dados de alta dimensionalidade e que possuem muitas características correlacionadas, em particular dados de imagens de faces corretamente detectadas e alinhadas;

- levantar hipóteses de como os métodos de origem remediam o problema especificado;

- propor novos métodos utilizando as hipóteses levantadas;

- validar as hipóteses através de experimentos com os métodos propostos.

## 1.6 Organização do texto

Na seção seguinte é proposta a metodologia de Agrupamento de Pixels para redução de dimensionalidade em imagens. A Seção 3 apresenta uma avaliação experimental das propostas de Agrupamento de Pixels para o reconhecimento de faces. A última seção apresenta considerações finais e proposta de trabalhos futuros.

## 2 Agrupamento de Pixels

Agrupamento de Pixels é uma metodologia para definição de métodos de extração de características. Utilizando Agrupamento de Pixels é possível definir vários métodos de extração de características. Esta metodologia segue o paradigma de agrupamento de características e inspira-se no fato de que uma imagem, em particular, imagens de faces humanas, tem várias regiões de intensidade semelhante. O objetivo da técnica é agrupar pixel para definir regiões na imagem, então extrair características destas regiões. Tais regiões não correspondem a uma simples segmentação de imagens aplicada a cada imagem da base individualmente. As regiões são as mesmas para todas as imagens da base.

A extração de características baseada em Agrupamento de Pixels depende de: (1) uma definição de representação para cada pixel de forma que esta representação corresponda a exatamente um pixel e represente este pixel em todas as imagens da base, (2) um algoritmo de agrupamento para agrupar as representações de pixels e (3) um método de extração de características para ser aplicado em cada região. Com essa metodologia proposta é possível definir novos métodos de extração de características, além de redefinir métodos já existentes, tais como *Waveletfaces* (Chien e Wu 2002) e *EigenSegments* (Shai Avidan 2002).

Considera-se que as imagens de face são imagens digitais representadas por uma matriz. Cada posição da matriz é um pixel, cujo valor indica a intensidade de iluminação em tons de cinza, indo do preto ao branco. As regiões extraídas possuem valor de intensidade semelhante dentro de uma imagem. O agrupamento de pixels define, de forma automática, regiões que possuem significado para os humanos: em imagens de faces é fácil identificar que estas regiões definem bochecha, testa, cabelo, barba, orelhas, pescoço, pano de fundo da imagem etc.

Para cada região é extraída uma característica. A partir desta característica é possível estimar o valor da intensidade dos pixels na região. Estimando-se a intensidade de todos os pixels de uma imagem, tem-se uma reconstrução da imagem. O erro de representação mede o quanto a imagem reconstruída está diferente da imagem original. Se as características extraídas dessas regiões minimizam o erro de representação é esperada uma boa reconstrução da imagem a partir das características extraídas.

As seções a seguir explicam com detalhes o método proposto. Na próxima seção é dada a definição de vetor-de-pixel, o elemento que irá ser agrupado para a construção das regiões. Em seguida, é explicado e exemplificado como se definir um método de extração de característica, utilizando a metodologia proposta. Um método de extração de características é proposto, utilizando a metodologia de Agrupamento de Pixels: Pedaços-por-Valor.

## 2.1 Vetor-de-pixel

O conceito principal do método proposto é a abordagem que forma grupos de pixels. Esta depende de como representar os pixels para realizar o algoritmo de agrupamento. Os objetivos são: obter uma partição que seja única para todas as imagens de treino; em cada imagem, os pixels de um grupo devem ter intensidades semelhantes. Para alcançar estes objetivos, foi criado o conceito de vetor-de-pixel. Cada região é a saída do algoritmo de agrupamento dos vetores-de-pixel extraídos do conjunto de treino.

Assumindo um conjunto de treino com $m$ imagens, em que cada imagem é uma matriz de dimensões $w \times h$, o número de pixel por imagem é $p = wh$. O **vetor-de-pixel** é o vetor que contém informação que é representativa para uma mesma posição de pixel em cada imagem do conjunto de treino. O vetor-de-pixel pode ser de dois tipos: (a) **vetor-de-pixel por valor**, um vetor $m$-dimensional que contém o valor do pixel na mesma posição para cada uma das $m$ imagens de treino; (b) **vetor-de-pixel por posição**, um vetor bidimensional que contém o valor da posição $(x, y)$ do pixel. Exemplos destes vetores-de-pixel estão descritos na Figura 1.

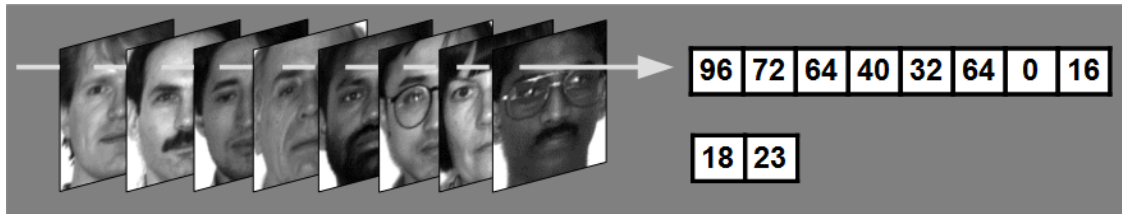

Figura 1: Exemplo de vetor-de-pixel por valor (acima) e vetor-de-pixel por posição (abaixo).

Nessa figura, o vetor-de-pixel por posição contém apenas as coordenadas do pixel. E o vetor-de-pixel por valor contém os valores de intensidade para cada uma das oito imagens do conjunto de treino. Estes valores de intensidade correspondem à mesma posição de pixel (18,23) em cada imagem. Em ambos os casos $p$ vetores-de-pixel são extraídos a partir de $m$ imagens de treino. É possível criar outras definições de vetores-de-pixel, mas este estudo restringe-se a estas duas. A seção seguinte explica como se emprega a metodologia de Agrupamento de Pixels para se construir uma técnica de extração de características.

## 2.2 Definindo um método de extração de características

A metodologia de Agrupamento de Pixels define um grupo de pixels como um conjunto não vazio de posições de pixels. Também assume que cada imagem de treino e teste tem exatamente o mesmo número de linhas e colunas, de forma que os mesmos grupos de pixels indexam qualquer imagem de treino ou teste. Uma nova característica extraída é uma função das intensidades dos pixels indexados por um grupo. Esta função pode ser uma transformação linear dos pixels indexados por um grupo, aqui chamada genericamente como uma projeção da imagem de entrada. O conjunto de vetores de projeção para cada grupo é a base do espaço de baixa dimensão onde ocorre a redução de dimensionalidade.

Para se definir um método de extração de características utilizando a metodologia de Agrupamento de Pixels é necessário: (a) uma definição de vetor-de-pixel; (b) um algoritmo de agrupamento e (c) uma função das intensidades dos pixels em uma região. Esta função pode ser uma combinação linear das intensidades de pixels de um grupo. Uma vez que esta configuração está definida, o algoritmo para extração de características tem três etapas principais:

1. extrair os vetores-de-pixel do conjunto de treino;

2. aplicar o algoritmo de agrupamento aos vetores-de-pixel;

3. criar, para cada grupo, uma projeção das suas características.

A primeira etapa é extrair $p$ vetores-de-pixel. Podem ser vetores-de-pixel por valor ou vetores-de-pixel por posição. A segunda etapa é agrupar os vetores-de-pixel através de um algoritmo de agrupamento, como o $k$-médias. Vale ressaltar que o número de vetores-de-pixel pode ser muito grande em problemas de alta dimensionalidade e que algoritmos de agrupamento aglomerativos, por exemplo, podem ser muito custosos no tempo. Dependendo do algoritmo de agrupamento escolhido, o resultado pode variar bastante formando diferentes regiões. A métrica de distância utilizada no agrupamento também pode influenciar a definição dos grupos finais. A etapa final consiste em extrair as características de cada grupo. Para extratores de características que serão definidos a seguir foi utilizada uma transformação linear. Um fluxograma para definição de método de extração de características com a metodologia proposta de agrupamento de pixels está descrito na Figura 2.

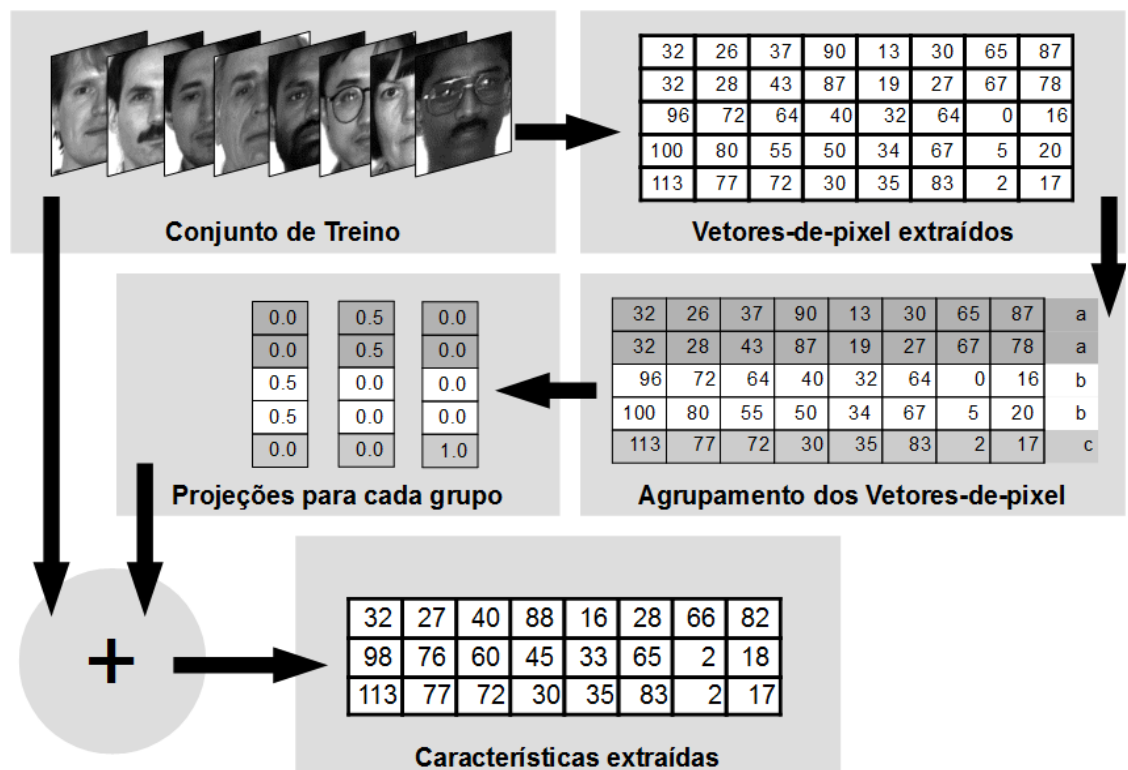

Figura 2: Fluxograma para definição de método de extração de características com a metodologia proposta de agrupamento de pixels.

Nessa figura é mostrado um conjunto de imagens de treino. A partir deste conjunto são extraídos vetores-de-pixel por valor. Cada vetor-de-pixel tem 8 dimensões pois o conjunto de treino tem oito imagens. Para simplificar o exemplo, são mostrados apenas 5 vetores-de-pixel como se cada imagem tivesse apenas 5 pixels. De fato, são calculados milhares vetores-de-pixel (um para cada pixel). Esse vetores-de-pixel são agrupados em três partições 'a', 'b' e 'c'. Para cada grupo é construída uma projeção, neste exemplo cada projeção corresponde à média dos pixels do grupo. Uma característica é extraída para a imagem como o produto interno entre o vetor que representa a imagem e o vetor que representa a projeção. Este produto interno equivale a projetar a imagem no vetor encontrado. As características extraídas estão representadas na figura. Na próxima seção é descrito um exemplo de um método de extração de características construído com a metodologia proposta.

## 2.3 Exemplo

Neste exemplo, é assumido um conjunto de treino de duas imagens 2×2, ou seja, $n = 2$ e $p = 4$. As colunas de cada imagem $X_i$ são empilhadas de forma que esta é convertida em um vetor $\mathbf{x}_i = [X_i(0,0), X_i(1,0), X_i(0,1), X_i(1,1)]^T$. As imagens $X_1$ e $X_2$ são convertidas em $\mathbf{x}_1 = [4, 1, 6, 3]^T$ e $\mathbf{x}_2 = [5, 2, 9, 7]^T$.

$$X_1 = \begin{bmatrix} 4 & 6 \\ 1 & 3 \end{bmatrix}, \quad X_2 = \begin{bmatrix} 5 & 9 \\ 2 & 7 \end{bmatrix}.$$

Para estas imagens, são criados quatro vetores-de-pixel por valor: $p_{(0,0)} = (4,5)$, $p_{(1,0)} = (1,2)$, $p_{(0,1)} = (6,9)$, e $p_{(1,1)} = (3,7)$.

Realiza-se o agrupamento destes vetores-de-pixels pelo algoritmo de agrupamento $k$-médias. Depois do agrupamento, dois grupos são criados $C_a = \{(0,0), (0,1), (1,1)\}$ e $C_b = \{(1,0)\}$. Para se extrair características, é necessário mapear cada agrupamento em cada imagem: os grupos para $\mathbf{x}_1$ são $(\{4,6,3\}, \{1\})$, para $\mathbf{x}_2$ são $(\{5,9,7\}, \{2\})$. Suponha que a regra de combinação para o terceiro passo seja $f(\mathbf{x}_i) =$ "escolha a média para cada grupo da imagem $\mathbf{x}_i$". De forma equivalente:

$$f(\mathbf{x}_i) = \begin{bmatrix} 1/3 & 0 & 1/3 & 1/3 \\ 0 & 1 & 0 & 0 \end{bmatrix} \mathbf{x}_i.$$

Então, as novas características para $\mathbf{x}_1$ e $\mathbf{x}_2$ são $f(\mathbf{x}_1) = (4,33; 1)$, $f(\mathbf{x}_2) = (7; 2)$, estas são as médias para cada grupo mapeado na imagem. As novas características para uma imagem de teste qualquer são calculdas a partir da mesma função, $f([1,2,3,4]^T) = (2,67; 2)$, na qual os grupos são $(\{1,2,4\}, \{2\})$. Na próxima seção é definida a técnica de extração de característica Pedaços-por-Valor (PV), funciona de forma similar a este exemplo.

## 2.4 Pedaços-por-Valor

Pedaços-por-Valor (PV) é um método de extração de características construído a partir da metodologia proposta de Agrupamento de Pixels. Em PV, vetores-de-pixel por valor são agrupados. Cada grupo define um conjunto de regiões da imagem. Espera-se que estas regiões tenham intensidade semelhante em qualquer imagem do conjunto de treinamento. Para cada grupo é extraída uma característica: a média aritmética das intensidades dos pixels que pertencem ao grupo. A seguir, uma definição formal de vetor-de-pixel por valor.

### 2.4.1 Vetor-de-pixel por valor

Um vetor-de-pixel por valor contém o valor para a mesma posição de pixel em todas as imagens do conjunto de treino. Assumindo que cada imagem no conjunto de treino tem o mesmo número de linhas e colunas, $w$ colunas e $h$ linhas, o número de pixels por imagem é $p = wh$. Então existem $p$ vetores-de-pixel. As amostras do conjunto de treino são imagens. Cada imagem é representada como um vetor coluna. Este vetor é o resultado de se empilhar cada coluna da matriz que representa a imagem.

Seja $\mathbf{x}_i$ a $i$-ésima amostra de treinamento

$$\mathbf{x}_i = [x_{i1}, \ldots, x_{ip}]^T, \tag{1}$$

$i = 1, \ldots, m$, então o $j$-ésimo vetor-de-pixel por valor, $j = 1, \ldots, p$ é

$$\mathbf{v}_j = [x_{1j}, \ldots, x_{mj}]^T, \tag{2}$$

em que $x_{ij}$ é a intensidade do $j$-ésimo pixel para a $i$-ésima imagem. Se existem $m$ imagens de treino, cada vetor-de-pixel tem $m$ dimensões.

### 2.4.2 Agrupamento

Seja $V$ o conjunto de todos os vetores-de-pixel extraídos do conjunto de treino,

$$V = \{\mathbf{v}_1, \ldots, \mathbf{v}_p\}, \tag{3}$$

um algoritmo de agrupamento rígido produz uma partição $L$ em $V$ com $n$ grupos,

$$L = \{V_1, \ldots, V_n\}, \tag{4}$$

de tal modo que $V_k$ é a $k$-ésima partição (pedaço), $k = 1, \ldots, n$; $V_k \subseteq V$, $V_k \neq \emptyset$, $V = \bigcup_{k=1}^n V_k$, e $V_k \cap V_l = \emptyset$ para $k \neq l$.

Cada partição é um grupo, um conjunto de vetores-de-pixel. Uma vez que cada vetor-de-pixel corresponde a uma posição na imagem, cada grupo é um conjunto de posições na imagem. Não existe a restrição de que um grupo deve formar uma região conexa. Para definir o agrupamento pode ser utilizado qualquer algoritmo de agrupamento rígido, nos experimentos da Seção 3 foi utilizado o $k$-médias.

### 2.4.3 Extração de Características

Dada qualquer imagem com as mesmas dimensões $w \times h$ das imagens de treino, uma característica é extraída para cada grupo. Isto se dá projetando a imagem $\mathbf{x}_i$ no espaço de características produzindo a nova característica $x'_{ik}$,

$$x'_{ik} = \mathbf{w}_k^T \mathbf{x}_i, \tag{5}$$

em que $\mathbf{w}_k$ é o vetor de projeção para a $k$-ésima característica extraída

$$\mathbf{w}_k = [w_{k1} \ldots w_{kp}]^T \tag{6}$$

e
$$w_{kj} = \begin{cases} 1/n_k & \text{if } \mathbf{v}_j \in V_k \\ 0 & \text{caso contrário} \end{cases}, \quad (7)$$

$n_k = |V_k|$ é o número de elementos em $V_k$. Deste modo a $k$-ésima característica extraída, $x'_{ik}$, é a média das intensidades dos pixels da $i$-ésima imagem cujas posições correspondem aos vetores-de-pixel no $k$-ésimo grupo.

A matriz de projeção $\mathbf{W}$ que tem a $k$-ésima linha igual a $\mathbf{w}_k^T$,

$$\mathbf{W} = \begin{bmatrix} \mathbf{w}_1^T \\ \vdots \\ \mathbf{w}_n^T \end{bmatrix}, \quad (8)$$

em que $n$ é o número de características extraídas.

O vetor de características extraídas $\mathbf{x}'_i$ de uma amostra de entrada $\mathbf{x}_i$ é calculado através da seguinte equação:

$$\mathbf{x}'_i = \mathbf{W}\mathbf{x}_i \quad (9)$$
$$= [x'_{i1}, \ldots, x'_{in}]^T, \quad (10)$$

em que $x'_{ik}$, $k = 1, \ldots, n$, é uma nova característica extraída, como na Equação 5. Note que o vetor de entrada $\mathbf{x}_i$ é $p$-dimensional, enquanto a imagem projetada $\mathbf{x}'_i$ é $n$-dimensional, $n \leq p$. De fato, para extração de característica em dados de alta dimensionalidade, como é o caso de imagens de face, é desejável que $n << p$.

### 2.4.4 Número de características extraídas igual ao número de grupos

São propostos dois casos quanto ao número de características extraídas. O primeiro extrai características para todos os grupos. O segundo extrai características para apenas alguns grupos, esta proposta está descrita na subseção seguinte. Em ambos os casos, apenas uma característica é extraída por grupo.

Nesta subseção considera-se o primeiro caso, isto é, todos os grupos são considerados. O número de características extraídas é um parâmetro do algoritmo de agrupamento, uma vez que uma característica é extraída para cada grupo. Em outras palavras, o número de características extraídas é igual ao número de grupos. Este é o caso de referência, considera-se que é extraída uma caracerística para cada grupo a não ser que especificado diferente.

### 2.4.5 Número de características extraídas menor que número de grupos

Na Análise dos Componentes Principais (PCA, *Principal Component Analysis*) é comum selecionar apenas as características projetadas que possuam a maior variância. A proposta desta subseção inspira-se no PCA, pois são extraídas mais características do que aquelas que serão utilizadas. Após a extração de características, são selecionadas as de maior variância e as de menor variância são descartadas.

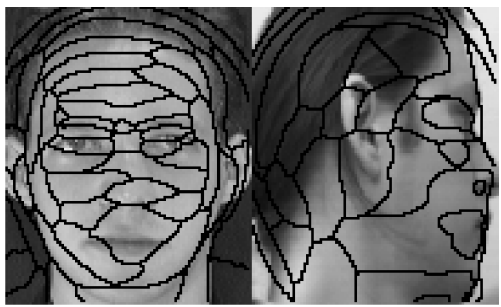

Figura 3: As regiões da imagens de faces são particionadas utilizando o método Pedaços-por-Valor para as bases ORL e UMIST, respectivamente.

Supõem-se que se pretende extrair $n$ características. Neste caso é necessário se formar $n+q$ grupos, $q = 1, 2, \ldots, (p-n)$. Então se extrai as características para os $n+q$ grupos no conjunto de treinamento. Depois calcula-se a variância para as $n+q$ características extraídas. Finalmente selecionam-se as $n$ características de maior variância.

### 2.4.6 Observações

É relevante notar que as características extraídas são muito dependentes do algoritmo de agrupamento. É este algoritmo o qual define que em um mesmo grupo estão os vetores-de-pixel que são mais similares. Para PV, um algoritmo de agrupamento rígido adequado é o $k$-médias, pois realiza em tempo hábil o agrupamento mesmo para um grande número de elementos. Para um número de exemplos muito grande, $k$-médias termina sua execução em menos tempo que outros algoritmos de agrupamento, por exemplo, agrupamento hierárquico aglomerativo.

É comum aplicar um algoritmo de agrupamento para os pixels de uma única imagem para se realizar segmentação de imagens, é o caso da técnica *superpixels* (Vargas et al. 2014; Rauber et al. 2013). Deve ser realçado que PV não é segmentação de imagem. O agrupamento não é realizado nos valores de pixel de uma única imagem mas é realizado no vetores-de-pixel (Seção 2.1). As posições dos pixels em um grupo definem uma região que não necessariamente é conexa. Deve ser enfatizado que esta região é a mesma para cada imagem.

Cada região tem propriedades similares dentro de cada imagem do conjunto de treino. É esperado que estas propriedades locais também ocorram nas amostras de teste. A média de cada região é o valor que minimiza o erro quadrático de representação. Como pode ser visto na Figura 3, essas regiões correspondem à localização dos olhos, nariz, sobrancelha, pescoço etc. Pode ser visto que para a base ORL, na qual as imagens estão de frente, que as regiões definem o contorno de áreas do rosto frontal. Na base UMIST, na qual a maioria das imagens está de lado, as regiões são distintas daquelas extraídas para a base ORL. Informações sobre estas bases estão disponíveis na Seção 3.1.

As três partes necessárias para se definir o método de extração de características Pedaços-por-Valor, utilizando as etapas descritas na Seção 2.2, são: (a) vetor-de-pixel por valor; (b) algoritmo de agrupamento rígido qualquer, em particular utiliza-se o $k$-médias com a distância Euclidiana é utilizado nos experimentos; (c) extração de características por região é a média da região pois minimiza o erro médio quadrático de representação, descrita acima como uma projeção sobre o vetor de projeção $\mathbf{w}_k$ da região $k$.

## 2.5 Conclusão

Esta seção apresenta a metodologia da Agrupamento de Pixels. Este método permite definir métodos de extração de características, com um viés para imagens de face. A metodologia baseia-se em encontrar regiões da face que possuam semelhança de acordo com um critério estabelecido. Estes critérios são definidos pelo conceito de vetor-de-pixel. Um extrator de características é definido utilizando a metodologia proposta: Pedaços-por-Valor (PV). Este agrupa pixels que têm intensidade semelhante e definem automaticamente regiões que possuem significado, tais como testa, sobrancelha etc. O método proposto extrai características como a média das intensidades dos pixels nas regiões formadas. A seção a seguir descreve experimentos utilizando os métodos propostos.

## 3 Experimentos com Agrupamento de Pixels

Esta seção apresenta experimentos para avaliar os métodos propostos sobre a metodologia de Agrupamento de Pixels. A técnica extração de características proposta é Pedaços-por-Valor. A Seção 3.1 descreve a metodologia dos experimentos. A Seção 3.2 realiza um experimento para definir um número adequado de grupos para a técnica Pedaços-por-Valor no problema de reconhecimento de faces. A Seção 3.3 explora a vantagem da técnica proposta de extrair características significantes, mesmo se posteriormente novas classes forem inseridas no problema. A Seção 3.4 analisa se é vantajoso para a classificação eliminar algumas das características extraídas. Finalmente, a Seção 3.5 discute os resultados dos experimentos.

## 3.1 Configuração para os experimentos

Os experimentos das seções seguintes foram realizados em duas base bem conhecidas: Yale e ORL, ambas utilizadas na seção anterior e em (Gao, Zhou e Pu 2013). As imagens de face Yale foram reduzidas para $92 \times 112$, que é o tamanho das imagens na base ORL. A base Yale tem 15 classes e 11 imagens por classe, as quais variam principalmente em condições de iluminação. A base ORL (também conhecida como base de faces AT&T) contém 40 classes e 10 imagens por classe que variam de forma suave tanto a pose quanto a iluminação. Além destas duas bases, foi utilizada a base UMIST. A base UMIST (também conhecida como base de faces Sheffield) tem um total de 574 imagens divididas de maneira não uniforme entre 20 classes. A base UMIST é utilizada em somente alguns experimentos, pois esta base é desbalanceada e possui muita varições de rotação da cabeça. As imagens desta base também foram re-escaladas para as dimensões $92 \times 112$. As bases ORL e Yale são mais adequadas para avaliar os métodos propostos.

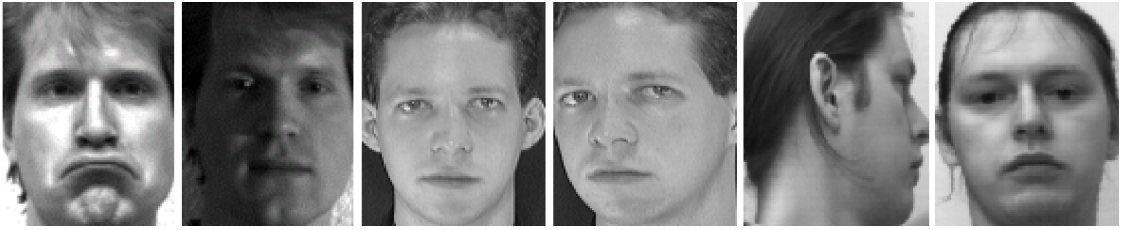

Figura 4: Duas imagens para cada base de faces, da esquerda para a direita: Yale e ORL.

A Figura 4 mostra alguns exemplos das imagens de cada base. Para estes experimentos os níveis de intensidade estão no intervalo [0; 1], isso significa que cada variável tem valor mínimo 0,0 e máximo 1,0.

A média da acurácia (Gao, Zhou e Pu 2013) para o classificador 1-NN foi calculada para repetições de experimentos do tipo *holdout*, a quantidade exata é especificada em cada caso. Em cada teste, metade das imagens da base foi escolhida como teste de forma estratificada e o restante das imagens foi utilizada para treino. Os mesmos conjuntos de treino e teste são utilizados em todos os métodos em uma execução de *holdout*. A taxa de acerto foi medida de 1 a 80 características extraídas. Uma vez que as imagens têm dimensões 92×112 o espaço original dos padrões é 10.304-dimensional.

Além do classificador 1-NN com distância Euclidiana, alguns experimentos utilizam os classificadores: SVM, Árvores de Decisão e Naive Bayes. Todos os quatro classificadores seguem as configurações padrões do Matlab 2013. A árvore de decisão está descrita em (Coppersmith, Hong e Hosking 1999), utilizando o índice Gini como critério para divisão, o mesmo índice utilizado no algoritmo CART. O SVM utiliza kernel linear e, para separar hiper planos, *Sequential Minimal Optimization* (SMO). O SVM utilizado é uma versão apenas para classificação binária. Para o caso onde existem múltiplas classes, é criado uma lista de classificadores em cascata. Um classificador diferencia a primeira classe das demais. Caso o exemplo não tenha sido classificado como sendo da primeira classe, é testado se é da segunda classe ou da terceira etc. O classificador Naive Bayes estima as funções de densidade utilizando um estimador não paramétrico com funções de *kernel* (Janela de Parzen), a função de *kernel* utilizada é uma função Gaussiana.

## 3.2 Escolhendo o número de características extraídas

Neste experimento é realizada a escolha do número $k$ de características extraídas para o método Pedaços-por-Valor. É verificada a acurácia média para três bases de imagens de face (ORL, UMIST e Yale) utilizando vários valores de $k$ (2, 4, 8, 16, 32, 64, 128, 256, 512, 1024 e 2048). Os resultados para a acurácia média de 10 *holdouts* está sumarizada na Figura 5. Em cada *holdout*, metade das imagens servem como conjunto de treino e as demais como conjunto de teste, foi utilizado o classificador 1-NN, conforme o o protocolo descrito na Seção 3.1.

Para a base UMIST, a acurácia máxima de 0,97 é atingida para 64 ou mais carac-terísticas extraídas. Para a base ORL, a acurácia máxima de 0,95 é atingida para 512 ou mais características extraídas. Para a base Yale, a acurácia máxima de 0,79 tambem é atingida para 512 ou mais características extraídas. Conclui-se que $k = 512$ é um número de características adequado para o reconhecimento de faces utilizando o método Pedaços-por-Valor nestas bases.

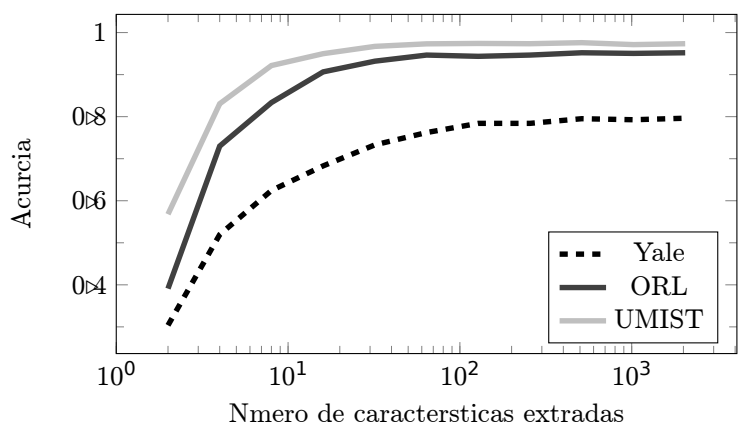

Figura 5: Acurácia média do método Pedaços-por-Valor para vários números de características extraídas. Em três bases de dados: Yale, ORL, e UMIST.

### 3.3 Inserindo novas classes

Este experimento avalia a robustez de um sistema de reconhecimento de faces, no caso em que são adicionadas pessoas novas ao sistema. Geralmente o classificador precisa ser treinado novamente, mas será que o método de extração de características precisa ser retreinado também? A avaliação aqui é realizada utilizando apenas poucas classes para treinar o método de extração de características. Então todas as classes são projetadas utilizando as projeções treinadas. São projetadas tanto as classes vistas no treinamento, quanto classes do conjunto de treino que não foram utilizadas para encontrar a projeção das imagens.

A acurácia é comparada para 512 características extraídas com Pedaços-por-Valor e para o máximo de características que podem extraídas com Autofaces. Autofaces cria no máximo tantas projeções quanto o número de amostras utilizadas para encontrar as projeções. Os resultados estão descritos na Tabela 1. Para cada base de dados (ORL, UMIST e Yale), foram realizados três experimentos: utilizando 1, 2 ou 3 classes escolhidas aleatoriamente.

Estes experimentos seguem o protocolo da Seção 3.1. A acurácia média e desvio padrão foram calculados para 10 *holdouts*. Em cada *holdout*, metade das imagens servem como conjunto de treino e as demais como conjunto de teste. Porém até três classes do conjunto de treino foram utilizadas para calcular as projeções utilizadas para reduzir a dimensão dos dados. Foi utilizado o classificador 1-NN nestes experimentos.

Treinar as projeções com apenas poucas classes têm um pequeno impacto na acurácia de Pedaços-por-Valor, mas não na acurácia de Autofaces. O método proposto tem acurácia de 11 a 23% maior que Autofaces para a base Yale, de 12 a 38% para a base ORL e de 3 a 10% para a base UMIST.

Tabela 1: Acurácia média e desvio padrão para Autofaces e Pedaços-por-Valor se apenas 1, 2, ou 3 classes são utilizadas para gerar as projeções. Classificador 1-NN, 10 repetições de *holdout* 50/50.

| # classes | Autofaces | Pedaços-por-Valor |
|---|---|---|
| YALE | | |
| 1 | 0,5280 ± 0,0651 | 0,7598 ± 0,0288 |
| 2 | 0,6378 ± 0,0390 | 0,7634 ± 0,0450 |
| 3 | 0,6756 ± 0,0252 | 0,7866 ± 0,0395 |
| ORL | | |
| 1 | 0,5595 ± 0,0623 | 0,9415 ± 0,0131 |
| 2 | 0,7715 ± 0,0515 | 0,9505 ± 0,0142 |
| 3 | 0,8245 ± 0,0369 | 0,9500 ± 0,0137 |
| UMIST | | |
| 1 | 0,8784 ± 0,0259 | 0,9721 ± 0,0079 |
| 2 | 0,9237 ± 0,0183 | 0,9746 ± 0,0099 |
| 3 | 0,9425 ± 0,0153 | 0,9735 ± 0,0087 |

A principal vantagem do método proposto é que este não tem a mesma limitação para o número de características extraídas que o Autofaces. Pedaços-por-Valor é adequado para extração de características para o reconhecimento de faces, uma vez que ele mostra vantagem significante se apenas poucas amostras são utilizadas para treinar as projeções. O método proposto também gera projeções discriminantes mesmo para classes que não foram vistas durante o treinamento.

### 3.4 Formando mais grupos do que o número de características

Este experimento explora a variante do método Pedaços-por-Valor que gera mais grupos do que o número de características extraídas. O objetivo deste experimento é verificar se este procedimento é vantajoso para a classificação. Este experimento utiliza o classificador 1-NN. É medida a acurácia média para 10 repetições de *holdout*, conforme o protocolo descrito na Seção 3.1. Foram utilizadas a bases de imagens de face Yale e ORL.

Neste experimento, os dados de treinamento são utilizados para definir três conjunto de regiões com: 64, 128 e 256 grupos. É calculada a acurácia para as características extraídas e removida a característica de menor variância. Este processo é repetido até que só exista uma única característica. Por exemplo, para o caso de 256 regiões é medida a acurácia utilizando 256 características de maior variância, 255 características de maior variância, e assim por diante até avaliar a acurácia utilizando apenas a característica de maior variância.

Os resultados dos experimentos estão descritos na Figura 6. Para a base Yale, percebe-se que a acurácia é maior para 64 características extraídas se forem gerados 128 ou 256 grupos e selecionadas novas variáveis de maior variância. O mesmo se dá para 128 características extraídas. A acurácia utilizando uma característica para cada um dos 128 grupos gerados é menor do que a acurácia selecionando as 128 características de maior variância para 256 grupos gerados. A acurácia máxima ocorre quando são utilizadas as 220 características de maior variância entre 256 característica extraídas.

Para a base ORL, não se percebe o mesmo fenômeno, a acurácia máxima é obtida quando nenhum grupo é desprezado. Nesta base, os resultados são opostos aos resultados obtidos para a base Yale. Nota-se a diferença dos resultados entre as duas bases pela posição das linhas no gráfico. Na base Yale, as linhas que representam menos grupos gerados ficam em baixo. Na base ORL, as linhas que representam menos grupos gerados ficam em cima.

Selecionar as características de maior variância pode melhorar a acurácia na classificação. Os resultados para esta técnica dependem da base de dados. Vale destacar que eliminar características extraídas equivale a ignorar regiões da imagem. Será vantajoso eliminar regiões se estas confundem o classificador. Uma análise mais detalhada deve verificar que tipos de regiões são estas. Podem ser regiões com problemas de iluminação, bem comuns na base Yale.

## 3.5 Discussão

Nesta seção, foram avaliados os métodos de Agrupamento de Pixels propostos. Na Seção 3.2 concluiu-se, para três bases de imagens de face, que 512 é um número de características adequado para o reconhecimento de faces com Pedaços-Por-Valor (PV). Na Seção 3.3, o método proposto mostrou-se muito robusto quando comparado ao PCA. Nesse experimento, apesar de serem utilizadas amostra de apenas poucas classes do conjunto de treino, não houve redução significativa na acurácia de PV. Na Seção 3.4, foi avaliado o efeito de não utilizar algumas características extraídas com PV. Concluiu-se que pode existir vantagem para o reconhecimento de faces, dependendo da base de dados. Os experimentos desta seção validam que a metodologia de Agrupamento de Pixel é útil para se definir métodos de extração de característica e compressão de imagens. Alguns experimentos indicam que os métodos propostos ainda podem ser aprimorados. A presente seção também pode inspirar novos métodos, derivados daqueles apresentados.

## 4 Conclusões

O Agrupamento de Pixels define uma estrutura chamada vetor-pixel. Esta estrutura pode ser de dois tipos. Um tipo contém apenas a posição do pixel na imagem. O outro tipo contém o valor de intensidade de um mesmo pixel para todas as imagens do conjunto de treinamento. A metodologia de Agrupamento de Pixels agrupa estes vetores definindo regiões na imagem e extrai uma característica como a média de uma região.

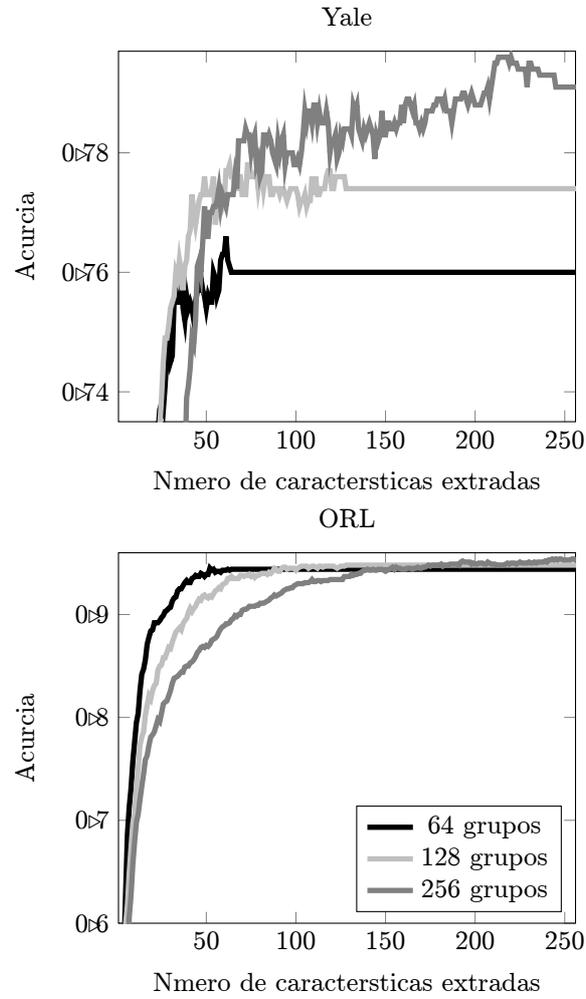

Figura 6: Acurácia média do método Pedaços-por-Valor para vários número de características extraídas. Em duas bases de dados: Yale e ORL. São gerados 64, 128 e 256 grupos, mas são selecionadas as características de maior variância.

Para as bases de imagens de face utilizadas, concluiu-se que as imagens podem ser suficientemente representadas para o reconhecimento por 512 atributos. Cada um destes atributos é a média de intensidade de uma região. O método proposto também mostrou-se robusto quando poucas imagens estão disponíveis para se encontrar a projeção.

Esta pesquisa, que é um recorte da minha tese "Agrupamento de Pixels e Autofaces Fracionário para Reconhecimento de Faces", começou ao se observar o método *Waveletfaces* (Chien e Wu 2002). Constatou-se que, para algumas bases de imagens de face, as imagens poderiam ser bastante reduzidas sem prejudicar o reconhecimento. Para tentar explicar este fato foi proposta a metodologia de Agrupamento de Pixels. Esta metodologia é utilizada para definir métodos que definem regiões para as imagens de face. Então uma característica é extraída para cada região. A seguir são apresentadas algumas propostas de trabalhos futuros.

## 4.1 Trabalhos Futuros

Para a técnica de Agrupamento de Pixels são propostos os seguintes trabalhos futuros:

- Pretende-se desenvolver novas definições de vetores-de-pixel e novos métodos de extração de características a partir dos grupos formados.

- Pretende-se desenvolver uma nova proposta para melhorar o reconhecimento através de um novo algoritmo de agrupamento. Percebeu-se que o $k$-médias, além de ser aleatório, produz grande variabilidade. Além disso, tem a desvantagem de poder gerar o caso em que um grupo pode esvaziar-se. A ideia de inserir apenas uma característica aleatória no grupo esvaziado não parece ser adequada, pois o algoritmo pode convergir com apenas uma característica no grupo. A proposta inicial é dividir o grupo de maior variabilidade em dois outros. Para se identificar o grupo de maior variabilidade basta somar os quadrados das distâncias de cada padrão ao centróide do seu novo grupo (que já são calculadas a cada iteração) e dividir pelo número de vetores-de-pixel no grupo. Isto não impactaria muito o custo de processamento do algoritmo. Para minimizar a variabilidade do método, pode-se iniciar os grupos com um conjunto de pixels adjacentes, por exemplo, uma região quadrada (um particionamento por grade).

- Pretende-se estender o método Pedaços-por-Valor combinando o algoritmo de agrupamento de características desenvolvido para o FAST (Song, Ni e Wang 2013) com o método de extração de características utilizados no Pedaços-por-Valor (a média do grupo) ou no AutoSegmentos (Shai Avidan 2002) (componentes principais do grupo). O FAST é uma abordagem de agrupamento de características que realiza apenas seleção de característica. Seus autores reconhecem que ele não é apropriado para o reconhecimento de faces. Realizar extração de características do grupo é uma boa forma de melhorar a aplicação do método para o reconhecimento de faces.

- Também pretende-se avaliar o método Pedaços-por-Valor para outros tipos de problemas que não reconhecimento de faces, utilizando bancos de dados de referência (*UCI Machine Learning Repository*).

- Além das modificações propostas acima, pretende-se verificar se é possível aproveitar ideias de outros métodos baseados em agrupamento de características, como (Song, Ni e Wang 2013; Jiang, Liou e Lee 2011; Sotoca e Pla 2010), para propor um método baseado em agrupamento de características mais adequado ao reconhecimento de faces.

- Para a aplicação de compressão de imagem, pretende-se definir um compressor para vídeo. A proposta consiste em dividir o video em conjuntos sequências, em que cada sequência possui muitos quadros com boa parte da cena em comum. Em seguida é gerado um modelo para cada sequência.

- Para compressão de imagens estáticas, pode-se dividir a imagem em várias regiões quadradas e criar um modelo para toda a imagem;

- Pode-se aproveitar este modelo para comprimir um conjunto de fotos que apresentam itens repetidos, por exemplo, pessoas, animais ou prédios.

- O algoritmo de compressão para imagens de faces pode ser estendido para gerar mais de um modelo por conjunto de faces. Por exemplo, um modelo para poses frontais e outros para poses laterais. Pode-se associar a pessoa primeiramente à pose do seu rosto para só então realizar o reconhecimento. Ainda pode-se tentar reconstruir uma imagem frontal a partir de uma imagem lateral e vice-versa, para só então realizar o reconhecimento. Como utilizar os modelos na classificação? Para cada amostra, verificar qual o modelo que obtém a menor média dos desvios padrão de cada grupo, este é o modelo adequado para a amostra. Podem existir alguns grupos de fundo os quais devem ser descartados em ambos os casos. Como gerar os modelos? A proposta é um algoritmo que agrupa ao mesmo tempo as amostras e as características. É um algoritmo iterativo de duas fases: 1, agrupa as amostras; 2, atualiza o modelo do grupo. O protótipo do grupo não é uma amostra mas um modelo, o exemplo é atribuído ao grupo do modelo para o qual obtém a menor média dos desvios padrões em cada grupo. Então é recalculado um modelo para cada grupo. É um algoritmo estilo $k$-médias, sendo que o protótipo é um modelo da face (um conjunto de regiões que são quase homogêneas). A medida de similaridade entre uma amostra e um grupo deve considerar para qual modelo há regiões mais homogêneas.

- Os modelos de regiões de face encontrados por Pedaços-por-Valor podem ser utilizados para comprimir vídeos de faces. Além disso, poderia ser considerado o seu uso para um padrão de transmissão para video telefonia.

- Extrair um particionamento dos autovetores do Autofaces. Atribuir cada característica ao grupo de índice igual ao do autovetor no qual ela tem o maior valor (não o maior valor absoluto, deve-se considerar o sinal). Comparar com Pedaços-por-Valor.

- Utilizar o método de agrupamento de características para identificar regiões simétricas da face e corrigir a iluminação de uma região com base na outra.

- BoW (*Bag of visual Words*) (Penatti et al. 2014) existem dois tipos de atribuição: rígido e suave. Na atribuição rígida cada característica local é atribuída a uma única palavra visual. Já na suave, cada característica é atribuída a todas as palavras visuais com um peso diferente para cada palavra. A atribuição suave possui alguma semelhança com as técnicas de agrupamento difuso, as quais poderiam ser utilizadas em algoritmos de agrupamento de características.

- Outra proposta é utilizar agrupamento de características na saída do BoW para tentar eliminar a redundância dos dados.

- Utilizar os algoritmos de agrupamento de *superpixels* (Vargas et al. 2014; Rauber et al. 2013) na técnica Pedaços-por-Valor.

- *Waveletfaces* é uma técnica que apresenta bom desempenho para o reconhecimento de faces (Chien e Wu 2002). *Waveletfaces* foi construído utilizando apenas a função Wavelet de Haar. Porém existe um conjunto de funções Wavelets infinito. É possível explorar o espaço das funções Wavelets, através de Wavelets parametrizáveis (Tewfik, Sinha e Jorgensen 1992). Esta proposta consiste em buscar uma função Wavelet que seja mais adequada para o reconhecimento de faces.

## Referências